\documentclass[11pt]{article}
\usepackage{acl2015}
\usepackage{times}
\usepackage{url}
\usepackage{latexsym}
\usepackage{amsmath}  
\usepackage{booktabs}  
\usepackage{makecell}  
\usepackage{graphicx,multirow}
\usepackage{orcidlink}
\usepackage[T1]{fontenc}
\usepackage[utf8]{inputenc}
\usepackage{textcomp}  

\title{VISTA: A Visual and Textual Attention Dataset for Interpreting Multimodal Models}

\author{
  Harshit~\orcidlink{0000-0002-7600-2219} \\
  School of Computing, \\
  University of Utah, \\
  Salt Lake City, UT, USA \\
  {\tt u1471783@umail.utah.edu} \\\And
  Tolga Tasdizen~\orcidlink{0000-0001-6574-0366} \\
  Scientific Computing and Imaging Institute, \\
  University of Utah, \\
  Salt Lake City, UT, USA \\
  {\tt tolga@sci.utah.edu} \\
}

\date{}

\begin{document}
\maketitle
\begin{abstract}
    The recent developments in deep learning (DL) led to the integration of natural language processing (NLP) with computer vision, resulting in powerful integrated Vision and Language Models (VLMs). Despite their remarkable capabilities, these models are frequently regarded as black boxes within the machine learning research community. This raises a critical question: which parts of an image correspond to specific segments of text, and how can we decipher these associations? Understanding these connections is essential for enhancing model transparency, interpretability, and trustworthiness. To answer this question, we present an image-text aligned human visual attention dataset that maps specific associations between image regions and corresponding text segments. We then compare the internal heatmaps generated by VL models with this dataset, allowing us to analyze and better understand the model's decision-making process. This approach aims to enhance model transparency, interpretability, and trustworthiness by providing insights into how these models align visual and linguistic information. We conducted a comprehensive study on text-guided visual saliency detection in these VL models. This study aims to understand how different models prioritize and focus on specific visual elements in response to corresponding text segments, providing deeper insights into their internal mechanisms and improving our ability to interpret their outputs.
\end{abstract}

\section{Introduction}

Vision-Language Models (VLMs) learn rich representations by leveraging both image and text modalities. They are capable of modeling the complex relationships between these two modalities, largely due to the vast and diverse datasets sourced from the internet on which they have been trained
\cite{lin2015microsoftcococommonobjects}
\cite{plummer2016flickr30kentitiescollectingregiontophrase}
\cite{schuhmann2022laion5bopenlargescaledataset} \cite{schuhmann2021laion400mopendatasetclipfiltered}. This extensive training enables VLMs to capture subtle connections between visual and linguistic information, allowing them to perform a wide range of tasks with impressive accuracy and generalization across various domains \cite{long2022visionandlanguagepretrainedmodelssurvey}
\cite{bugliarello2021multimodalpretrainingunmaskedmetaanalysis}
\cite{Mogadala_2021}.

\begin{figure}[!t]
\centering
\includegraphics[width=2.8in]{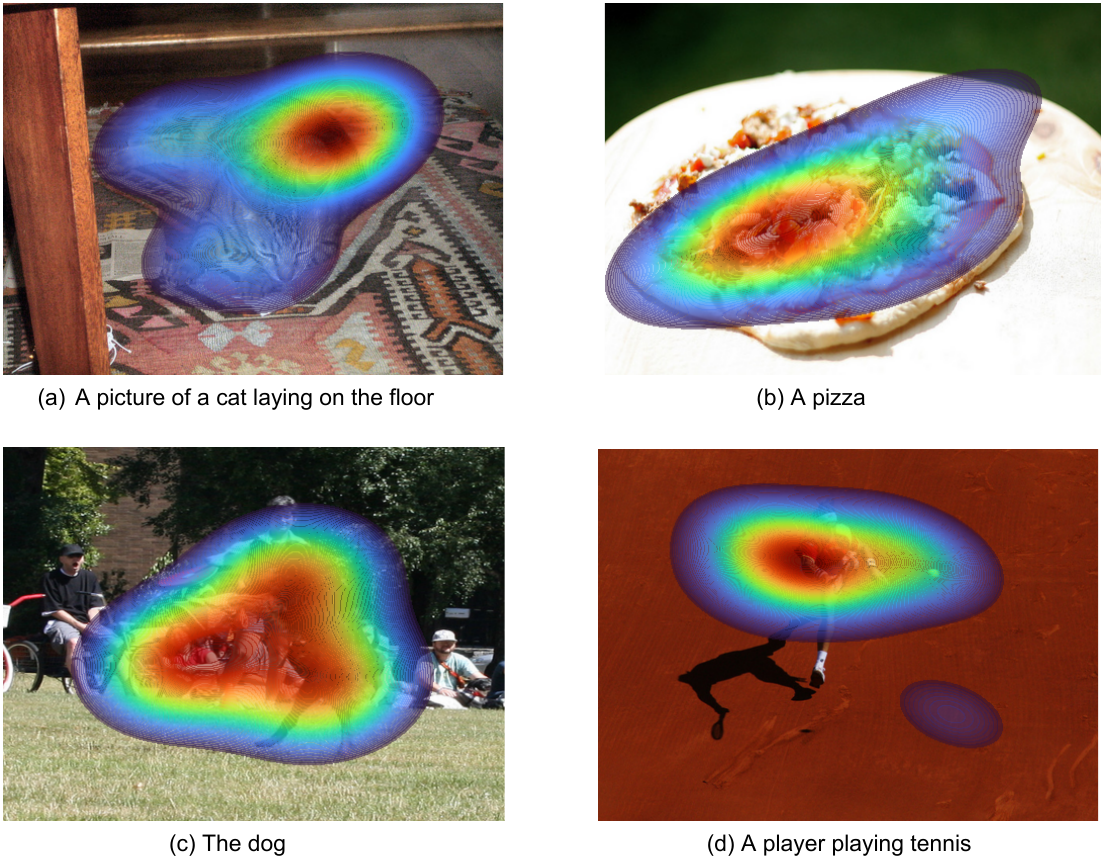}
\vspace{-1em}
\caption{Example of heatmaps from VISTA with their corresponding texts.}
\label{heat}
\vspace{-1em}
\end{figure}

Despite their impressive performance, VLMs, like many other deep learning models, often lack interpretability \cite{yue2023mmmu}. This lack of transparency has prompted researchers to probe into the inner workings of these VL models to understand the nature of the representations they learn \cite{lei2023licoexplainablemodelslanguageimage} \cite{bousselham2024legradexplainabilitymethodvision}
\cite{stan2024lvlminterpretinterpretabilitytoollarge}. By investigating how these models process and integrate visual and textual information, researchers aim to uncover the underlying mechanisms that drive their decision-making processes and enhance their interpretability. Various interpretability techniques, such as Grad-CAM \cite{Selvaraju_2019} and spatial attention maps, generate saliency maps that highlight the spatial regions most significant for each of the network’s outputs. We propose to measure the similarity between human attention and the saliency maps produced by these methods. This comparison can also serve as a tool for validating and refining these interpretability techniques, ensuring that the models are not only accurate but also aligned with human understanding in critical applications. 

\section{Related Work}
\textbf{Vision-Language Models (VLMs)}: In recent years, substantial progress has been made in research on integrating language and vision. Numerous tasks now combine language—ranging from words, phrases, and sentences to paragraphs and full documents—with visual data, typically in the form of images or videos. Initially, much of the work focused on linking low-level linguistic units, such as words, with images or videos to create visual-semantic embeddings (Barnard et al., 2003; Frome et al., 2013; Kiros et al., 2014b; Liu et al., 2015; Cao et al., 2016; Tsai et al., 2017; Guo et al., 2018; Mogadala et al., 2018b; Wang et al., 2019; Kim et al., 2020). These embeddings are valuable for various downstream applications, as well as for understanding adversarial attacks (Wu et al., 2019) to enhance model robustness.

More recently, leading research labs have consistently introduced cutting-edge VLMs, such as OpenAI's CLIP \cite{radford2021learningtransferablevisualmodels}, Salesforce's BLIP \cite{li2022blipbootstrappinglanguageimagepretraining}, and DeepMind's Flamingo \cite{alayrac2022flamingovisuallanguagemodel}. Notable examples like GPT-4 \cite{openai2024gpt4technicalreport} and Gemini \cite{geminiteam2024geminifamilyhighlycapable} highlight the ongoing evolution of chatbots within the VLM space. However, not all multimodal models are VLMs; for instance, text-to-image models like DALL-E \cite{ramesh2021zeroshottexttoimagegeneration}, Stable Diffusion \cite{rombach2022highresolutionimagesynthesislatent} and Midjourney (https://www.midjourney.com/) do not have a language generation component, illustrating the varied landscape of multimodal AI. A typical VLM architecture includes separate image and text encoders to generate embeddings, which are then combined in an image-text fusion layer, and the fused vector is processed by a large language model (LLM) to produce the final visually-aware text output.

\textbf{Eye-tracking Saliency Map Datasets}: Numerous eye-tracking databases have been developed to study and model visual attention behavior. One such large-scale saliency dataset is MIT1003 \cite{5459462}, consisting of 1003 images sourced from Flickr and LabelMe. Widely used benchmark databases include MIT300 \cite{Judd_2012} and CAT2000 \cite{borji2015cat2000largescalefixation}, containing 300 and 2000 test images, respectively. SALICON \cite{7298710}, the largest crowd-sourced saliency dataset, comprises 10,000 training images, 5,000 validation images, and 5,000 test images, and is commonly used for pretraining saliency prediction models.  
In addition to the existing image saliency datasets, only one image-text saliency dataset has been developed. \cite{lanfredi2023comparingradiologistsgazesaliency} dataset utilizes eye-tracking (ET) data from five radiologists, offering valuable insights into visual attention in medical imaging. The study found that interpretability maps generated from multiple chest X-ray (CXR) classification models could be as similar to the radiologists' ET maps as those generated by other radiologists.

\section{VISTA: Visual-Textual Attention Saliency Dataset}


We manually curated a dataset for human image-text alignment with the help of a group of human annotators. The data collection process involved presenting volunteers with images and asking them to describe the scenes depicted in each image. Throughout this process, we recorded the volunteers' eye movements to capture their visual attention patterns, as well as their verbal descriptions via voice recordings. These voice recordings were later transcribed to text, and the original audio files were deleted to ensure de-identification and protect the privacy of the participants. This careful approach to data collection not only preserved the integrity of the visual attention data but also ensured compliance with ethical standards regarding participant confidentiality. The resulting dataset provides a rich resource for understanding the relationship between visual attention and linguistic descriptions, offering valuable insights for advancing interpretability in vision-language models.

After gathering the eye movement data and corresponding textual descriptions, we synchronized both datasets using time stamps. During our experiment, we used the EyeLink 1000 Plus \cite{eye_l} eye tracker to accurately record participants' eye movements as they viewed and described images. We adopted the eye-tracking setup from \cite{eye_l} recommendations. In this setup, each image was viewed by a single participant during the experiment. After completing the recordings, the initial results contained some noise, as anticipated. To address this, we carefully selected the image-text-saliency triplets that exhibited the closest alignment. This process resulted in a final dataset of 508 well-aligned image-text saliency maps. Although this approach reduced the total amount of data, it ensured high-quality results, resulting in a robust dataset.

\section{Metrics and baselines}
Eye-tracking (ET) maps were created by applying Kernel Density Estimation (KDE), with the density estimate at each point weighted according to the duration of each eye fixation. This method ensures that regions, where participants focused their gaze for longer periods, are given greater emphasis in the resulting map. By assigning more weight to longer fixations, the KDE captures not just the frequency of eye movements, but also the intensity of attention directed towards specific areas within the visual field. This approach provides a more nuanced representation of visual attention patterns, highlighting areas of cognitive interest or importance in a way that simple fixation counting methods might overlook. Figure ~\ref{heat} shows some of the examples. The resulting ET maps offer insights into how participants distribute their visual focus over time, enabling a deeper understanding of attentional behavior.

Based on the existing literature on automatic generation of human saliency maps, we selected two metrics for comparing saliency maps: normalized cross-correlation (NCC), also known as Pearson’s correlation coefficient, and the Borji formulation of the area under the curve (AUC). The NCC was computed directly between the eye-tracking (ET) map and the generated saliency map. This metric quantifies the similarity between the two maps by measuring their correlation over the entire image. Specifically, NCC calculates how closely the patterns in both signals (saliency maps) match, providing an overall assessment of their alignment.
The NCC uses the following,

\begin{align*}
\text{NCC}(x_1, x_2) &= \frac{1}{P-1} \sum \left( \frac{(x_1(p) - \mu_{x_1})}{\sigma_{x_1}} \right. \\
&\quad \left. \times \frac{(x_2(p) - \mu_{x_2})}{\sigma_{x_2}} \right)
\end{align*}


\noindent where $x_1$ and $x_2$ are the two images of same size, $x_i(p)$ is the value of image $x_i$ in pixel coordinate $p$, $P$ is the total number of pixels in one image, $\mu_{x_i}$ is the average value of $x_i$, and $\sigma_{x_i}$ is the unbiased standard deviation of $x_i$.

The NCC measures the degree of linear correlation between the two images, with values ranging from -1 to 1, where 1 indicates perfect positive correlation, 0 indicates no correlation, and -1 indicates perfect negative correlation. This method provides a robust assessment of the global similarity between the ET map and the generated saliency map, as it accounts for the overall distribution of pixel intensities across the image.

\begin{table*}[t]
\caption{Scores for Image-Text Alignment and Open-Vocab Segmentation Models}
\label{tab:combined_scores}
\centering
\begin{small}
\begin{tabular}{l@{\hskip1em}l@{\hskip1em}c@{\hskip1em}c}
\toprule
  & Model & NCC & AUC \\
\midrule
\multirow{5}{*}{\parbox{2.0cm}{\centering Image-Text \\ Matching Models}}
 & CLIP & 0.13 [0.14,0.14,0.12,0.13,0.14] & 0.57 [0.58,0.58,0.57,0.57,0.58] \\
& BLIP-ITM-Base & \textbf{0.24} [0.25,0.23,0.24,0.25,0.25] & 0.63 [0.64,0.63,0.63,0.63,0.63] \\
& BLIP-ITM-Large & 0.17 [0.18,0.17,0.17,0.16,0.18] & 0.60 [0.60,0.60,0.60,0.59,0.61] \\
& ALBEF & 0.19 [0.20,0.19,0.21,0.18,0.19] & 0.57 [0.58,0.57,0.58,0.57,0.57] \\
& ViLT & -0.02 [-0.02,-0.01,0.00,-0.03,-0.015] & 0.49 [0.49,0.50,0.50,0.48,0.49] \\
\midrule
\multirow{4}{*}{\parbox{2.0cm}{\centering Open-Vocab \\ Saliency Models}}
 & CLIP-Seg & \textbf{0.31} [0.33,0.31,0.31,0.30,0.31] & \textbf{0.67} [0.68,0.67,0.66,0.66,0.67] \\
& OV-Seg & 0.18 [0.17,0.18,0.16,0.18,0.17] & 0.59 [0.59,0.59,0.58,0.59,0.59] \\
& OpenSeg & 0.14 [0.15,0.14,0.13,0.14,0.15] & 0.58 [0.58,0.57,0.57,0.58,0.58] \\
& ODISE & 0.16 [0.17,0.17,0.18,0.16,0.16] & 0.59 [0.60,0.59,0.60,0.59,0.59] \\
\bottomrule
\end{tabular}
\end{small}
\label{res}
\end{table*}

The AUC (Area Under the Curve) metric evaluates the performance of a saliency map by treating each pixel as the output of a binary classifier tasked with determining whether the pixel corresponds to a human fixation or not. In this context, the classifier is expected to assign higher values to locations where human fixations occurred (positive locations, derived from the eye-tracking map) and lower values to other locations (negative locations, represented by a uniform heatmap). The AUC provides a measure of how well the saliency map distinguishes between these two types of locations.

To compute the AUC, we sample from the heatmaps, which represent the spatial probability distributions of visual attention. Specifically, the metric assesses the likelihood that the classifier assigns a higher value to a positive location (where a fixation occurred) than to a negative one (where it did not). The closer the AUC is to 1, the better the saliency map is at predicting human fixations; a score of 0.5 would indicate random guessing, while values below 0.5 suggest the map is performing worse than chance.

Through experimentation, we varied the number of samples to determine the impact on the accuracy of the AUC calculation. By analyzing the standard deviation of AUC scores across different sample sizes, we found that using 1000 positive samples and 1000 negative samples provided a reliable and sufficiently accurate estimation of the AUC. This approach balances computational efficiency with the need for precise evaluation, ensuring that the metric reflects the true performance of the saliency map without introducing unnecessary noise or variability.

People often apply center bias adjustments when recalculating metrics like NCC or AUC to account for the natural tendency of humans to focus their gaze toward the center of an image. However, after visual inspection of the dataset, we did not incorporate such bias adjustments in our analysis, as our dataset does not exhibit a center bias. This decision allows for a more direct and authentic evaluation of the saliency map's ability to capture genuine areas of visual attention as dictated by the participants' eye movements, rather than artificially emphasizing central regions of the image.


\section{Results and Discussion}
Table \ref{res} presents the results of evaluating multiple models on VISTA. Initially, we conducted standard evaluations on the full dataset to measure each model's performance. To further ensure the reliability of our findings, we employed a bootstrapping technique, which involved randomly sampling the dataset with replacement over five iterations. The corresponding results are indicated in brackets. Bootstrapping provides a means to estimate the stability and consistency of our metrics by creating and evaluating multiple subsamples, thereby offering a more robust assessment of model performance.

We first conducted experiments on image-text alignment models by visualizing the attention maps produced by various Vision and Language Models (VLMs). The goal was to investigate the extent to which these models accurately align the attention weights with the human eye-tracking (ET) data in our dataset. The VLMs used for this task included CLIP \cite{radford2021learningtransferablevisualmodels}, ViLT \cite{vilt}, BLIP \cite{blip}, and ALBEF \cite{albef} all of which were pre-trained on large-scale datasets and fine-tuned for the task of image-text alignment.

\textbf{CLIP} \cite{radford2021learningtransferablevisualmodels} performed moderately on both metrics, with an NCC score of 0.13 and an AUC of 0.57. While these results indicate some alignment with human attention patterns, they suggest that CLIP struggles to capture finer details of human visual focus, as evidenced by the relatively low NCC score.

\textbf{BLIP-Base} \cite{blip} trained for image-text matching was one of the top-performing models in this evaluation, achieving an NCC of 0.24 and an AUC of 0.63. These results suggest that BLIP-ITM-Base aligns more closely with human visual attention compared to other models, indicating that it may better capture the relationships between visual regions and corresponding text. Despite its larger architecture, BLIP-Large trained for image-text matching underperformed compared to the base model, with an NCC of 0.17 and an AUC of 0.60. This drop in performance might indicate that the larger model overfits to certain aspects of the training data, reducing its ability to generalize well to human-like attention patterns.

\textbf{ALBEF} \cite{albef} showed reasonable performance with an NCC of 0.19 and an AUC of 0.57. Its performance was comparable to CLIP, indicating moderate alignment with human visual attention but with room for improvement in capturing nuanced attention cues.

\textbf{ViLT} \cite{vilt} exhibited the weakest performance in our experiments, with a negative NCC score (-0.02) and an AUC of 0.49, suggesting that the model failed to align with human attention patterns. The negative NCC indicates that the model's attention maps were not only uncorrelated but possibly misaligned with human attention, highlighting its limited interpretability in this context.

Next, we evaluated text guided image segmentation model. The task involved providing a textual description of an image and analyzing how well the model's segmentation map captured the relevant areas described in the text. For example, when the text mentioned "a red car," the segmentation map was expected to highlight the red car in the image. We compared the generated segmentation maps to the human ET maps to assess the degree of alignment.

 

Here, the results were mixed. CLIP-Seg \cite{lüddecke2022imagesegmentationusingtext} showed the strongest performance among the tested methods, with an NCC of 0.31 and an AUC of 0.67. These scores suggest that CLIP-Seg aligns relatively well with human attention and is effective at detecting salient regions in an open-vocabulary context. It outperforms other methods, especially in terms of NCC, indicating its potential for tasks that require accurate saliency prediction. OV-Seg \cite{liang2023openvocabularysemanticsegmentationmaskadapted} performed moderately, achieving an NCC of 0.18 and an AUC of 0.59. Although it does not match the performance of CLIP-Seg, it still demonstrates reasonable alignment with human attention. This suggests that OV-Seg is somewhat effective at identifying salient regions but may lack the finer precision required for more accurate saliency detection. OpenSeg \cite{ghiasi2022scalingopenvocabularyimagesegmentation} delivered the lowest performance with an NCC of 0.14 and an AUC of 0.58. These results indicate that the model struggles to align well with human visual attention, likely due to less effective saliency detection mechanisms. ODISE \cite{xu2023openvocabularypanopticsegmentationtexttoimage} achieved an NCC of 0.16 and an AUC of 0.59, performing similarly to OV-Seg. While its AUC score shows some capacity to capture salient regions, the NCC indicates a relatively weak correlation with human attention, suggesting room for improvement in its saliency detection capabilities.





\section{Conclusion}

In this work, we introduced \textit{VISTA}, a human-annotated visual and textual attention dataset, to explore and enhance the interpretability of Vision-Language Models (VLMs). By aligning eye-tracking data with textual descriptions, our dataset provides a unique perspective on how humans associate visual regions with corresponding text segments. Through the evaluation of multiple VLMs using well-established metrics like NCC and AUC, we demonstrated varying degrees of alignment between human attention and model-generated saliency maps, with models such as BLIP-ITM-Base and CLIP-Seg showing promising results. However, our results also highlight the challenges that VLMs face in capturing nuanced human visual attention, particularly in complex tasks like image-text alignment and segmentation.

The findings underscore the importance of human-centric datasets like VISTA in advancing the interpretability and transparency of VLMs. By providing insights into the internal mechanisms of these models, this work paves the way for future research aimed at improving the reliability and trustworthiness of multimodal systems. Furthermore, our dataset and methodologies serve as valuable tools for developing more human-aligned interpretability techniques, ultimately contributing to safer and more explainable AI systems in vision and language applications.

\bibliographystyle{acl}
\bibliography{references}


\end{document}